# Integrating Transit Signal Priority into Multi-Agent Reinforcement Learning based Traffic Signal Control

**Dickness Kakitahi Kwesiga**
School of Civil and Environmental Engineering
Georgia Institute of Technology
Atlanta, GA 30332, USA
Tel: 404-423-4022 Email: dkwesiga3@gatech.edu

**Suyash Chandra Vishnoi, Ph.D.**
School of Civil and Environmental Engineering
Georgia Institute of Technology
Atlanta, GA 30332, USA
Tel: 512-785-7634 Email: scvishnoi@gatech.edu

**Angshuman Guin, PhD.**
School of Civil and Environmental Engineering
Georgia Institute of Technology
Atlanta, GA 30332, USA
Tel: 404-894-5830 Email: angshuman.guin@gatech.edu

**Michael Hunter, PhD.**
School of Civil and Environmental Engineering
Georgia Institute of Technology
Atlanta, GA 30332, USA
Tel: 404-385-1243 Email: michael.hunter@ce.gatech.edu

Word count: 7476 words

Submission Date: August 1, 2024

# ABSTRACT

This study integrates Transit Signal Priority (TSP) into multi-agent reinforcement learning (MARL) based traffic signal control. The first part of the study develops adaptive signal control based on MARL for a pair of coordinated intersections in a microscopic simulation environment. The two agents, one for each intersection, are centrally trained using a value decomposition network (VDN) architecture. The trained agents show slightly better performance compared to coordinated actuated signal control based on overall intersection delay at v/c of 0.95. In the second part of the study the trained signal control agents are used as background signal controllers while developing event-based TSP agents. In one variation, independent TSP agents are formulated and trained under a decentralized training and decentralized execution (DTDE) framework to implement TSP at each intersection. In the second variation, the two TSP agents are centrally trained under a centralized training and decentralized execution (CTDE) framework and VDN architecture to select and implement coordinated TSP strategies across the two intersections. In both cases the agents converge to the same bus delay value, but independent agents show high instability throughout the training process. For the test runs, the two independent agents reduce bus delay across the two intersections by 22% compared to the no TSP case while the coordinated TSP agents achieve 27% delay reduction. In both cases, there is only a slight increase in delay for a majority of the side street movements.

# 1 INTRODUCTION

Transit Signal priority (TSP) involves providing preferential treatment to transit vehicles at signalized intersections. Previous TSP research is primarily focused on developing optimal and more adaptive algorithms to improve bus performance while limiting impacts to general traffic. More recent studies integrate some level of connected vehicle (CV) technologies (1-12) in TSP algorithms. Adaptive TSP algorithms have mainly been based on mathematical programming (2, 10, 13-17) but recent efforts are testing machine learning approaches, especially reinforcement learning (RL) (18-23). These ongoing efforts build on recent and still evolving RL-based adaptive traffic control algorithms (24-33). RL-TSP formulations have mainly considered single RL agents with only a few recent attempts considering multiple cooperating agents (21, 34, 35).

The few studies that extend RL-based control algorithms to integrate TSP do so by modifying state and reward functions to include transit flow parameters. However, bus arrivals at intersections are sparse occurrences compared to general vehicles. This means that training the same RL agent to learn general traffic control and bus prioritization may take prohibitively longer periods, especially in higher fidelity traffic microscopic simulation environments. To overcome this, several studies inflate the bus arrivals to generate sufficient training samples for bus control (19, 20, 22, 34). The robustness of these models to generalization to low frequency bus arrivals and the associated impacts on general traffic are not reported. It is likely that these models may unnecessarily penalize nonpriority movements in cases of low frequency buses schedules.

Considering the above stated gaps, in a previous study the authors formulated and trained two separate agents, one for background traffic control, and one for TSP (36), in a single intersection application. The TSP agent is formulated as an event-based algorithm that is only activated when a transit vehicle reaches within the defined connected vehicle (CV) communication range of the intersection. In the absence of a bus, the intersection runs on the general traffic control agent.

The current effort extends the single agent formulation in the previous study to a multi-agent framework. The previous study considered a single isolated intersection without the need to consider coordination, platooned arrivals, midblock bus stops, scalability issues, and the interaction of different agents in the same environment. The first part of the current study formulates multi-agent reinforcement learning (MARL) based adaptive signal control algorithm for a pair of intersections, not in the presence of a transit vehicle. Two agents, one controlling each intersection are trained under centralized training and decentralized execution paradigm (CTDE). The second part of the study formulates event triggered RL agents to implement TSP when a transit vehicle reaches within the defined CV communication range of the intersection. Two variations of MARL TSP agents are considered (1) independent TSP agents with decentralized training decentralized execution (DTDE) and (2) coordinated TSP agents under CTDE framework. Moving from single agent to multiple agents required consideration of new challenges including equilibrium selection, scalability, non-stationarity, and partial observability of the environment. The agents are trained and tested in a traffic microscopic simulation environment with architecture decisions made to improve run time efficiencies that becomes more critical when moving from single agent to multiple agents.

# 2 RELATED STUDIES
## 2.1 RL for Traffic Signal Control

   (a) Single Agent RL for traffic signal control

Several studies have formulated and tested RL-based adaptive traffic signal control algorithms in simulation environments. These studies show that RL-based algorithms can potentially outperform the conventional fixed and actuated signal timings (24-33). Most of these studies formulate and train RL control agents based on deep q-network (DQN) and its variations, while a few have utilized actor-critic algorithm and its variations. Agent actions mainly involve selecting the next phase to assign green time from a discrete set of available phases. States are defined with data available from traffic sensors and from connected vehicles (CVs). Rewards functions are formulated based on the common traffic flow performance metrics, such as delay, waiting time, queuing, number of stops and throughput.

(b) Multi-agent RL for traffic signal control

Arterial signal timing optimization can be readily formulated as a fully cooperative MARL problem with a single agent controlling each intersection and cooperating with the adjacent intersection agents to generate a coordinated signal timing plan. A few recent efforts have attempted to extend RL-based traffic control from a single intersection to multiple intersections and networks (37-42). Several of these efforts are focused on using CTDE to achieve signal coordination through cooperation between individual intersection agents, with extensions to explicit communication modules to exchange observation and action histories across agents (37, 39-41). Others utilized independent q-learning (IQL) (38, 40). Despite the environment non-stationarity challenges of IQL the studies report improvements in traffic flow performance compared to fixed signal timing. As indicated by Rashid et al. (43), when trained long enough, IQL individual agents may be able to select global optimum actions and IQL can serve as a good baseline for more sophisticated MARL algorithms.

## 2.2 Single Agent RL-based TSP

A few recent studies have attempted to integrate TSP in adaptive RL-based traffic control algorithms (18-23). TSP integration is mainly achieved by modifying state and reward functions to include bus flow parameters and performance metrics. Entries for bus flow parameters are filled when the bus is in the network, otherwise they are set to zero or left blank. Weighting factors are utilized to define the level of desired priority for buses compared to other traffic. These factors are commonly based on vehicle occupancy and bus schedule deviations.

## 2.3 Multi-intersection/coordinated transit signal priority

A few research efforts have been undertaken to develop algorithms to implement coordinated TSP strategies across multiple intersections. The majority of these studies are based on mathematical programming (2, 10, 13-15). Recent studies by Li et al. (35) and Long and Chung (34) apply multi-agent RL to implement coordinated TSP across multiple intersections.

(a) Mathematical programming approaches

Zeng et al. (10) formulates an online mixed integer linear program (MILP) to determine optimal signal timings at any two consecutive intersections on the arterial. MILP algorithms formulated by Hu et al. (2) and Truong et al. (14) determine the optimal green reallocation signal timings at downstream intersections within a given threshold distance. Ma et al. (15) formulated a mathematical programming-based model to select coordinated TSP strategies for intersections falling within adjacent bus stops. Zhai et al. (13) formulates a MILP to select coordinated TSP strategies across multiple intersections with the underlying aim of minimizing bus bunching. All these studies reported improvements when TSP strategies are coordinated across intersections compared to when independent TSP strategies are selected for each intersection.

**(b) RL-based approaches**

Long and Chung (44) extends the single agent RL algorithm developed in Long et al. (22) to MARL to implement TSP cooperatively across multiple intersections. The study utilizes value decomposition network (VDN), a CTDE formulated by Sunehag et al. (45). The agents are trained to minimize person-delay in the network. In the performance evaluation, the fixed and variable phase sequence RL implementations are compared with a coordinated fixed timing plan. Li et al. (35) formulates and tests a DQN based centralized training and centralized execution (CTCE) algorithm to implement TSP for self-driving buses across two intersections. The study claims superior performance of the developed TSP algorithm compared to Conventional TSP (CTSP). However, side street impacts are not addressed in the paper. Additionally, CTCE is known to suffer from scalability issues.

Overall, there are still several gaps in RL-based traffic signal control research and more so in RL-based TSP. Transit performance objectives are often in conflict with general traffic performance objectives which needs to be considered in RL reward formulations. Bus arrivals at intersections are sparse occurrences compared to general traffic which makes it challenging to train a single agent to control general traffic and transit buses optimally. Event-based TSP agents may be more appropriate. There is very limited literature on MARL TSP control.

# 3   METHODOLOGY

## 3.1   Multi agent reinforcement learning and VDN Overview

### 3.1.1   Multi agent reinforcement learning

MARL involves multiple agents interacting with a common environment and with each other. Moving from single agent to multiple agents introduces new challenges that require modification or formulation of new learning frameworks. These challenges include equilibrium selection, non-stationarity and partial observability of the environment, scalability, and multi-agent credit assignment (46, 47). MARL problems are normally formulated as partially observable Markov decision processes (PoMDP). Considering the stated challenges of MARL, different training and execution modes have been formulated to address the challenges. These include decentralized training decentralized execution (DTDE), centralized training and centralized execution (CTCE), and centralized training decentralized execution (CTDE).

In DTDE, each agent uses local observation and action histories to learn a local policy. The trained local policy is executed conditioning on local observations. This is also called independent RL, a common example being independent q-learning (IQL). The main challenge of DTDE is the non-stationarity of the environment due to the actions of other agents. Despite this challenge, several studies have reported good results with DTDE and it has been used as a reasonable baseline to evaluate the performance of more advanced algorithms (43). In CTCE, some type of information or mechanism is centrally shared both in training and execution. At the top of this category is where the agents are reduced to a single agent using global observation and action histories to learn a global policy that is executed centrally. The key challenge of CTCE is scalability as the global and action spaces grow exponentially with the number of agents (48). Considering the challenges of purely DTDE and CTCE algorithms, there emerged a new paradigm of CTDE algorithms that is a hybrid of the DTDE and CTCE. In CTDE, agents share some mechanism or information during training, but the learned policies are executed locally conditioning on local observations. Two CTDE algorithms, value decomposition networks (VDN) formulated by Sunehag et al. (45) and Qmix formulated by Rashid et al. (43) are increasingly popular in recent MARL studies. Both VDN and Qmix learn a central value action function with the two distinguished by how the global value action function is decomposed. This paper adopts VDN architecture. VDN is chosen over Qmix for its simpler architecture and is able to achieve good performance, as shall be seen in the results section.

### 3.1.2 Value Decomposition Network (VDN)

As shown in Equation 1, the joint value function $Q_{tot}$ is formulated as the sum of individual agent value functions $Q_i$ which condition on local observations $h^i$.

$$Q_{tot} = \sum_{i=1}^{d} Q_i \left( h^i, a^i \right) \tag{1}$$

Complete architecture of implementation is described in the Sunehag et al. (45) paper but the key processes are as follows. On the forward pass, the network for each agent takes its local observations as the input and outputs the local value function. The value functions are summed up to generate $Q_{tot}$, which together with the common reward are used to compute the loss function. The local action values ($Q_i$) are learned by backpropagating the gradients. During execution, local actions are predicted by the learned action value function conditioning only on local observations.

### 3.2 Test network

As shown in Figure 1, the network model consists of two intersections A and B, 1600 ft apart. The 1600ft spacing is chosen to have the two intersections where coordination is likely beneficial while limiting the possibility of queue spillback. The main street is four lanes (two lanes each direction) with both intersections having a left turn bay on their EB and WB approaches. All N/S minor streets are two lanes (one lane each direction) with left turn bays on all approaches. The speed limit on both the main and cross street is assumed to be 40 mph and in Vissim the desired speed of all vehicles is set to 40 mph. Vissim default parameters for car following and lane changing models are maintained as this network is not modeled based on a field network. At both intersections, the main street has 1440 veh/h and 171 veh/h for through and left turns, respectively, while the cross street has 270 veh/h and 257 veh/h for through and left turns, respectively. For simplicity, no vehicles are assumed to turn right. The bus route modeled is Eastbound on the main street. There is a far side bus stop on the route downstream of intersection A. The communication range is taken as 800 ft, typical of DSRC, although cellular or other technology could be utilized. The simulation time step is 0.1 seconds for all experiments.

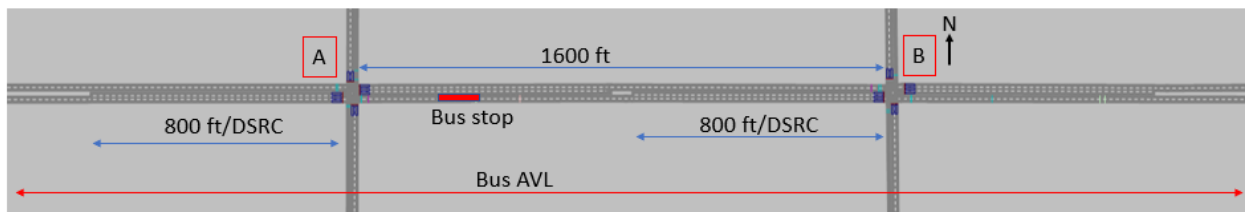

Figure 1: Network Model in Vissim

Bus movement is modeled using PT line and PT line stop modules available in Vissim. Dwell time at the bus stop is modelled based on actual field Automated Passenger Counted (APC) data obtained from Metropolitan Atlanta Rapid Transit Authority (MARTA). In Vissim, dwell time is entered as a continuous distribution function (CDF). Vissim assigns a dwell time to each bus using the stop by randomly sampling from the entered dwell time distribution. When the assigned dwell time is zero, the stop is skipped.

## 3.3   Model formulation

### 3.3.1   Architecture and Experiment design

For the background controller (signal control when a transit vehicle is not present) an adaptive MARL algorithm based on VDN is formulated, trained, and tested. Control switches to the transit control agent when a transit vehicle is present. For transit control two TSP MARL frameworks are formulated (1) independent TSP agents with DTDE and (2) coordinated TSP agents with CTDE based on VDN architecture. TSP agents are formulated as event-based algorithms triggered only when a transit bus reaches within the zone of the subject intersection or upstream intersection in case of coordinated TSP. All agents formulated in this study have a double deep q-network (DDQN) architecture in which there is a main network for predicting q-values and a target network for predicting the target q-values. The details of DDQN can be found in (49).

The first part of the study formulates the adaptive MARL algorithm for traffic control and compares this against coordinated actuated signal control. In the network shown in Figure 1, coordination is in the eastbound (EB) direction, which is also the direction of the bus route. Coordination for general traffic along the bus direction enables testing the effectiveness of TSP when the bus falls out of the platoon to serve passengers at the midblock bus stop. Each intersection is controlled by a single RL agent with the agents trained together with two objectives of (1) minimize total intersection delay and (2) prioritize coordination in the EB direction (see section 3.3.2). Note that with only the first objective, the agents may reach a two-way coordination that minimizes total traffic delay. The second objective is added considering the subsequent TSP experiments.

The second part of the study uses the trained RL agents as background traffic controllers to train two TSP agents, one for each intersection (see section 3.3.4). As already indicated two variations are considered, (1) independent TSP agents in DTDE framework and (2) coordinated TSP agents under CTDE framework. In the independent TSP variation, the objective is to train an agent at each intersection to implement TSP by selecting locally optimum TSP strategies. Only the intersection with the bus on the approach activates the TSP agent while the other intersection runs on the background controller. In the coordinated TSP variation, it is hypothesized that selecting coordinated TSP strategies improves networkwide bus performance and limits networkwide impacts to the cross-street traffic. The two centrally trained TSP agents become active when the bus enters the zone of the upstream intersection. The coordinated TSP algorithm is based on VDN architecture as in the background MARL signal control algorithm.

### 3.3.2   Definition of state, action and reward for general traffic control agents
####   (a) State

The problem is formulated as PoMDP in which each intersection agent has only access to local observations at that intersection. The study assumes availability of CV data which enables more detailed definition of the traffic state. At each intersection the state is defined by two components: (1) vehicle state ($Veh_{state}$) and signal state ($Sig_{state}$). As shown in Equation 2, $Veh_{state}$ is a vector of the number of vehicles (v) in each lane (n) on each approach (m). The study assumes knowledge of the number of vehicles within the communications zone. In the signal state definition, the phase receiving green, yellow, or red clearance is populated with the elapsed duration of the indication. All other entries are set to zero. Whereas the vehicle state is defined locally for each intersection, it is assumed that each intersection agent has access to the signal data for the adjacent intersection. Thus, the signal state definition for each agent consists of the signal state at both intersections. This is illustrated with the two columns of $Sig_{state}$ in Equation 3. In extending the model to more than two intersections, each agent would have access to signal state at only its immediate neighboring intersections.

$$Veh_{state} = \begin{bmatrix} v_{11} \\ v_{12} \\ . \\ . \\ v_{mn} \end{bmatrix} \quad (2)$$

$$Sig_{state} = \begin{bmatrix} 0 & 12 \\ 22 & . \\ 0 & 0 \\ 0 & 0 \end{bmatrix} \quad (3)$$

### (b) Action

Action in this study is defined as the selection of the next phase from the available phases at that time step. This may be either keeping the current phase or selecting a different phase. For simplicity, at each intersection four phases are defined, North/South Left, North/South Through, East/West Left, and East/West Through. On both the main and side streets, left turn movements start and end at the same time and through movements start and end at the same time.

A key challenge of training RL agents in traffic microscopic simulation is run time efficiency, which worsens moving from single agent to multiple agents. Shorter RL time steps mean more frequent interruption of the simulation and thus higher run times. As shall be seen in the results, hundreds or even thousands of simulation runs were needed to reach convergence of each model. A base RL time step of 1 second is defined but optimized depending on the current phase and duration at the two intersections. For instance, if both intersections are serving yellow or minimum green of the respective current phase, the time step is taken as the minimum of the remaining minimum green at both intersections.

At every time step, a set of valid actions is determined, and invalid action masking (IAM) algorithm formalized by Huang and Ontañón (50) is used to mask out the invalid actions in the action selection algorithm. Valid actions are selected considering the constraints of minimum green, maximum green, clearance phases and phasing sequence. In the implementation of IAM algorithm, predicted q-values of invalid actions are replaced with large negative numbers to prohibit the selection of the corresponding actions. Cycle length is allowed to vary from cycle to cycle.

### (c) Reward

For centralized training, a reward function is selected to optimize the overall network traffic performance. The global reward, R was formulated as the average of local reward, $r_1$ at the first intersection and local reward $r_2$ at the second intersection, R= mean($r_1$,$r_2$). The local reward formulation is shown in Equation 4. The base local reward is taken as the average of the delay ($d_i$) for all vehicles (n) at the intersection. Two negative penalty terms are imposed on the local reward. The first penalty N is imposed when queue length on any of the side street movements ($ql_j$) exceeds a set threshold ($ql_{Th1}$). The penalty is aimed at preventing excessive queuing on the side streets especially in the early training stages. The second penalty M is imposed when the phase changes from $\phi_t$ to $\phi_{t+1}$ before the end of maximum green when the queue length on the current phase $ql_{\phi t}$ exceeds a set threshold $ql_{Th2}$. This penalty is aimed at reducing instability in earlier training stages. Adding the two penalty terms significantly sped up the convergence. It was also desired to have coordination prioritized in the direction of the bus movement (EB). This was achieved by including an extra term in the reward function. The base offset ($\theta$) is estimated considering the block length and free flow speed. As a form of initialization, the two agents receive a bonus positive reward (P) if the achieved offset ($\theta'$) was within a set $\Delta\theta$ seconds of the base offset ($\theta$).

$$reward\ r = -\frac{\sum_1^n d_i}{n} - N - M + P \qquad (4)$$

$$N = \begin{cases} 0, & ql_j \leq ql_{Th1} \\ -9999, & ql_j > ql_{Th1} \end{cases}$$

$$M = \begin{cases} 0, & \\ -9999, & \phi_{t+1} \neq \phi_t, ql_{\phi t} > ql_{Th2} \end{cases}$$

$$P = \begin{cases} 0, & \\ 100, & \theta' - \Delta\theta \leq \theta \leq \theta' + \Delta\theta \end{cases}$$

### 3.3.3  Model training

After formulating the models, simulation runs are performed in Vissim, archiving training data at each time step that is later retrieved for training at the end of each episode (i.e., simulation run). Each episode lasts for 30 minutes with the same random seed utilized for all runs. Event based scripts are preferred to drive the Vissim simulation as opposed to COM for their better runtime efficiency as detailed by Kwesiga et al. (36). Simulation runs and training are performed on an x64-based PC equipped with 12th Gen Intel(R) Core i9-12900, 2400 MHZ, 16 Core(s), 24 Logical Processor(s),128 GB of RAM, Intel (R) UHD Graphics 770 GPU and Windows 11 operating system.

Preliminary training is undertaken to select the model hyperparameters. For brevity, the results of hyperparameter tuning are not included in the paper. The final selected hyperparameters include: 3 layers of 128, 256 and 256 neurons with rectified linear unit activation function for both main and target networks, learning rate = 0.01, optimizer = Adam, discount rate (gamma) = 0.99, exploration probability decay rate = 0.01, memory buffer capacity = 20000, and target network update frequency = 50 episodes.

### 3.3.4  Definition of state, action and reward for TSP agents

The state definition for general traffic is expanded to include bus flow parameters. It is assumed that all buses are connected, broadcasting their location and speed utilizing basic safety messages (BSM) as they approach the intersection. The communication zone of the approach link is divided into 25 feet cells. The 25 ft cell size is selected to ensure that there is at most one bus in each cell at any given time step. If a bus occupies two cells its position is set to the most downstream cell. Two vectors are created to represent bus position and speed as shown in Equation 5 and Equation 6 respectively. When the bus occupies a cell, the corresponding entry in the location matrix is populated with 1 and the with bus speed in the speed vector, otherwise both entries are populated with zero. Equations 5 and 6 show an example of the bus in the 4th cell from the beginning of the link traveling towards the stop line with a speed of 36 mph.

$$Bus_{pos} = [0, 0, 0, 1, 0, 0 \ldots \ldots, 0]$$ (5)

$$Bus_{speed} = [0, 0, 0, 36, 0, 0 \ldots \ldots, 0]$$ (6)

**(a) Action**

Action for TSP agents is also the selection of the next phase in the variable phasing sequence which consists of the same discrete set of entries as in the general traffic controller discussion. IAM algorithm is applied to mask out the invalid actions.

**(b) Reward**

The objective of the TSP algorithm is to minimize bus delay and number of stops at the intersections while limiting adverse impacts to the side street. For independent agents, the local reward function is formulated as a combination of bus delay ($b_d$), bus speed ($b_v$), and a penalty term for queuing on the side street as shown in Equation 7. The side street queuing penalty (M) is as described earlier, imposed when queue length on any of the side street movements exceeds a set threshold.

$$reward\ r = \begin{cases} -b_d + b_v & ql_j \leq ql_{Th1} \\ -b_d + b_v - M, & ql_j > ql_{Th1} \end{cases}$$ (7)

For coordinated TSP, where the two TSP agents are trained together, the local reward at each intersection is formulated as a function of general traffic delay, bus delay, and bus speed. As shown in Equation 8, weighting factors are imposed on bus delay ($W_{bd}$) and bus speed ($W_{bs}$) indicating the desired level of priority. A penalty, M is imposed to prevent excessive queuing on the side street. When the bus is on intersection A, bus delay and speed in the local reward for intersection B are zero and vice versa. The global reward is taken as the average of the two local rewards (R= mean($r_1$,$r_2$).

$$reward\ r = \begin{cases} -\dfrac{\sum_1^n d_i}{n} - W_{bd}b_d + W_{bv}b_v, & ql_j \leq ql_{Th1} \\ -\dfrac{\sum_1^n d_i}{n} - W_{bd}b_d + W_{bv}b_v - M, & ql_j > ql_{Th1} \end{cases}$$ (8)

# 4 RESULTS

This section discusses the study results. The first part compares the VDN based MARL for general traffic control with a coordinated actuated signal system under the same traffic conditions. The second part presents and discusses the performance of the event-based TSP control algorithms working on top of the VDN based MARL.

## 4.1 Performance of VDN based MARL

Figure 2 shows the VDN based MARL learning curve plotting episode number on horizontal axis and average reward on the vertical axis. Average reward is obtained by averaging rewards obtained at each time step within the episode. Each episode lasts for 30 minutes, with training data sampled after the run completion. As the agents are centrally trained the learning curve is common to both agents. The reward is a global reward obtained by averaging the two local rewards, one from each agent. The agents progressively learn the policy, converging after approximately 900 episodes. The learning process also exhibits relatively good stability.

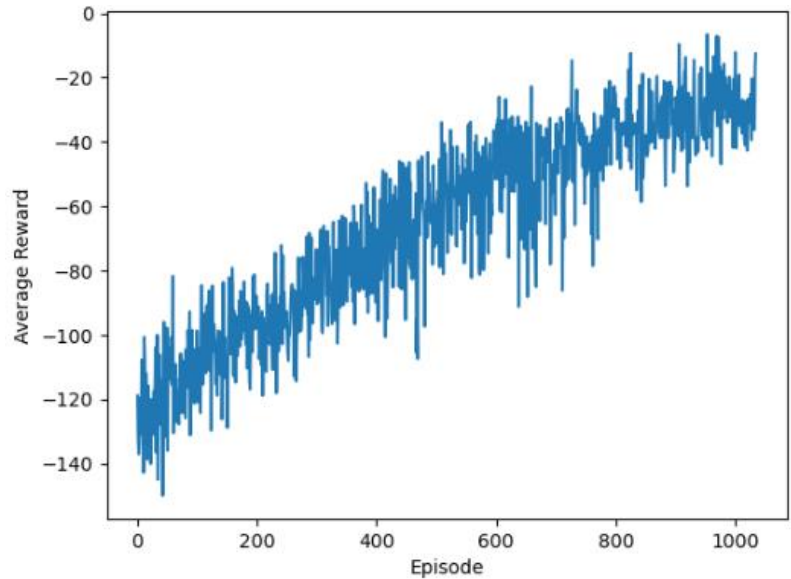

Figure 2: VDN Based MARL learning curve

Figure 3 shows a comparison of the performance of trained VDN based MARL and coordinated actuated signal control (ASC) at a volume to capacity ratio (v/c) of 0.95. Coordinated ASC timing parameters are determined offline using commonly available commercial software. To allow for fair comparison with the VDN MARL solution similar constraints are imposed on the coordinated ASC solution, limiting to paired left turns and paired through movements on both main and cross streets. This effectively reduces the timing to four-phase single ring control. Within Figure 3 six origin-destination (OD) movements are shown including corridor eastbound through (EB_TH), corridor westbound through (WB_TH), intersection A southbound through (A_SB_TH), intersection A northbound through (A_NB_TH), intersection B southbound through (B_SB_TH), and intersection B northbound through (B_NB_TH). EB_TH is the coordinated movement. Each box plot includes data from 10 replicate runs with different random seeds. Each run lasts for 1 hour with the first 15 minutes as the warmup period with the below results over the last 45 minutes. The red square is the average of the data while the extents of the boxes indicate the upper and low quartiles.

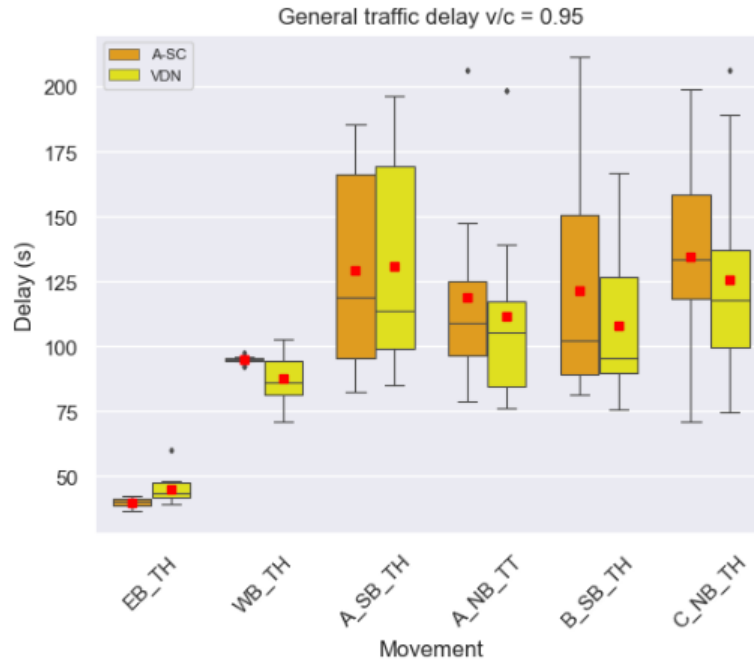

Figure 3: Comparing VDN MARL with ASC at v/c=0.95

ASC shows slightly better performance for the coordinated movement (EB_TH) but slightly worse performance for the opposite direction (WB_TH). This is an indication that the VDN solution is providing slightly more weight to the WB movement while the ASC solution more heavily favors the EB movement, although both solutions are achieving superior EB coordination. On the side street both control systems show high variability. VDN based MARL shows slightly lower means and medians in all but one instance, as well as lower range values. Overall VDN based MARL arguably shows slightly better performance compared to ASC. As shown by Kwesiga et al. (36), the superiority of the RL based control is likely to be more pronounced at lower v/c values where increased flexibility in the optimization with more frequent gap out opportunities.

It is seen that the implicit communication in VDN and the selected reward formulation achieved good coordination between the two signals without additional explicit communication modules as applied by Chang et al. (37), Bokade et al. (40), Chen et al. (41). The training achieves a key objective of having an adaptive control algorithm working at least as well as ASC and achieving progression in the EB direction for the subsequent TSP experiments with the bus route in the EB direction.

## 4.2    Independent TSP agents

### (a)    Learning rate and convergence

Figure 4 shows bus delay evolution during training for the two intersections. Each episode lasts for four hours with average bus headways of 15 minutes. The same random seed is utilized across the episodes. Each plotted point represents the average delay for all buses within each episode, which is typically 16 buses per episode. For both agents, bus delay progressively reduces as the agents learn the optimal policy, and converges after approximately 200 episodes. However, sizable variability in bus delay persists even after convergence, which is indicative of the presence of instability in the training. This instability associated with DTDE is not unexpected as both agents interact with Vissim at the same time, leading to non-stationarity of the environment.

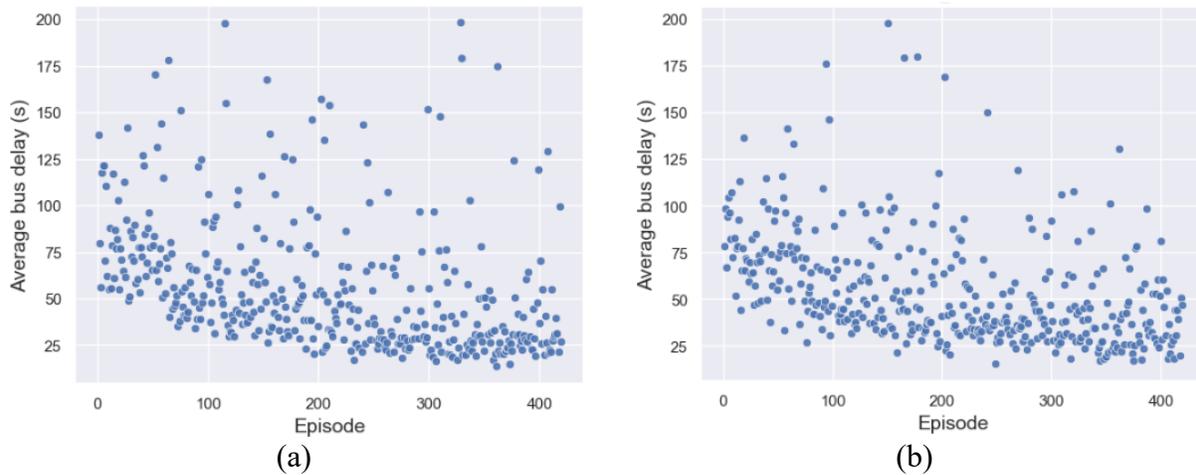

Figure 4: Average bus delay at (a) Intersection A and (b) Intersection B during training for independent TSP agents

### (b) Bus travel time

New simulation runs are performed to test the trained TSP agents. Each run lasts for 4 hours and includes 16 buses. Average bus headways are 15 minutes (900 seconds) with a random term added to the headway to generate random arrival profiles. Figure 5 shows bus travel time, with and without TSP, measured across three sections: (1) crossing intersection A, ending prior to the far side bus stop, labeled Inter A_EB; (2) crossing intersection B, starting immediately after the bus stop. labeled Inter B_EB; and (3) crossing intersections A and B, labeled Inter A&B_EB. Inter A&B_EB includes dwell time at the bus stop. The Figure 5 plots are based on the average travel time value from each of the 10 simulation runs. The no TSP case measures bus travel time with the two intersections running on VDN based MARL with no special bus treatment. The TSP case measures travel time with the TSP agents activated when the bus approaches the intersection. It is seen that TSP significantly reduces bus travel time across both intersections. The high variability of travel time in the no TSP case across intersection B (i.e., Inter_B_EB) is explained by the bus occasionally falling out of the platoon to serve bus passengers at the stop. TSP significantly reduces this travel time variability. Averaging across runs, the travel time across both intersections reduces by 22% with TSP activated.

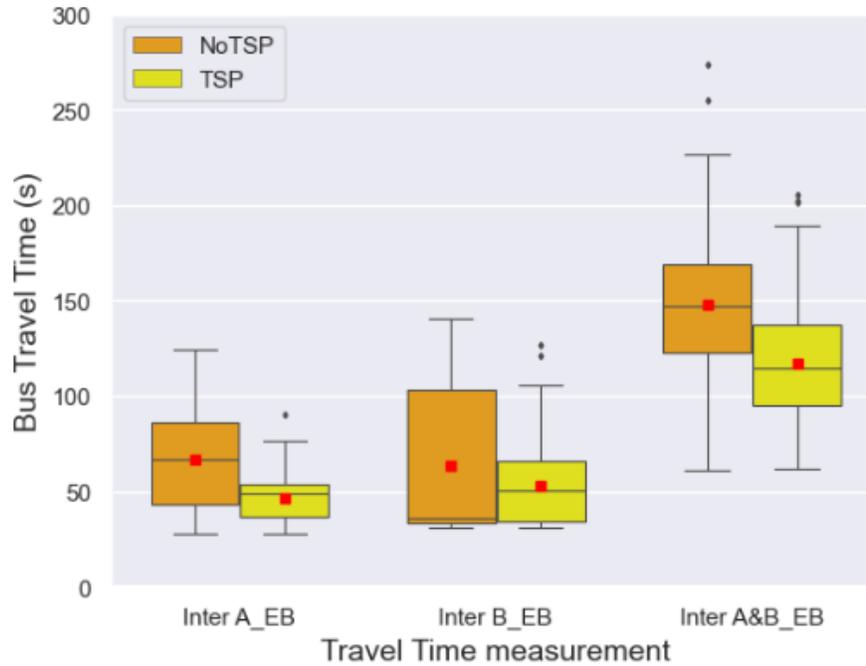

Figure 5: Bus travel time with and without TSP for Independent TSP agents

### (c) Side street Delay

Figure 6 shows general traffic delay for selected side street movements with and without TSP for averaged across the ten simulation runs. A_SB_TH and A_SB_LT stand for the southbound through and left-turn movements at intersection A, respectively, while B_SB_TH and B_SB_LT stand for the southbound through and left-turn movements at intersection B, respectively. The vehicles included in the analysis traverse the intersection in the time interval of five minutes (300 seconds) after bus check-in. The interval of 300 seconds is chosen following a study by Guin et al. (51) that showed that for v/c of 0.95, side street delay change persists up to approximately 300 seconds. For all movements except for B_SB_LT, general traffic delay only slightly increases when TSP is activated. The adaptive algorithm is able to successfully adjust the side street movement phasing, compensating for the awarded bus priority. For B_SB_LT delay increases by about 18% after TSP. Additional refinement to the reward function could be considered to better balance the impacts across movements.

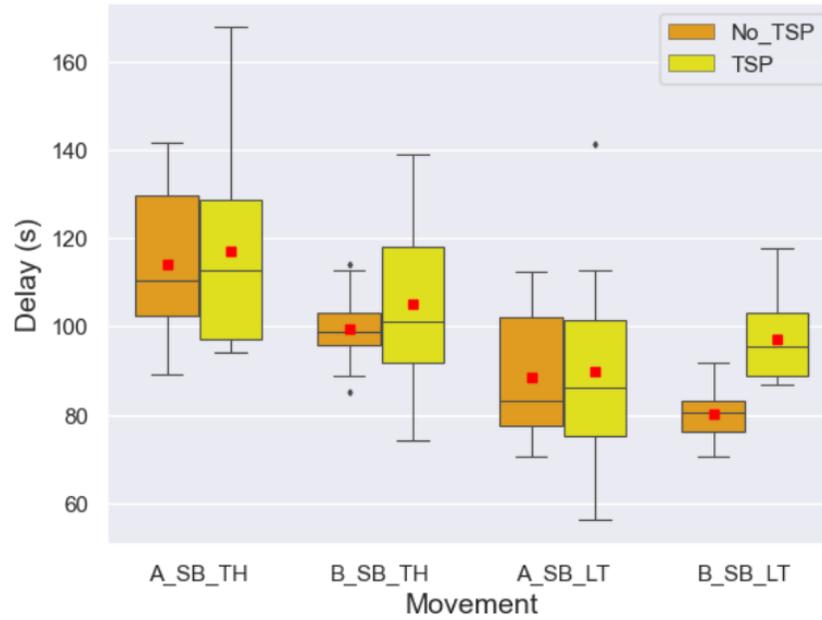

Figure 6: Side Street delay with and without TSP

## 4.3 Coordinated TSP agents

### (a) Learning rate and convergence

Figure 7 shows bus delay evolution during training for (a) intersection A and (b) intersection B for coordinated TSP agents. Different from the independent TSP experiment, in this case, each episode lasts for 1 hour with average bus headways of 15 minutes. The same random seed is utilized across episodes. Each plotted point is the average delay for all buses within the episode, which is typically 4 buses. For both agents, bus delay progressively reduces as the agents learn the optimal policy, converging after approximately 800 episodes. Compared to independent agents, the training exhibits good stability, which is facilitated by centralized training. A comparison of Figure 4 and Figure 7 illustrates the challenge of the non-stationarity associated with DTDE. Nevertheless, in both cases of independent training and centralized training, the agents converge to the same bus delay value of approximately 25 seconds.

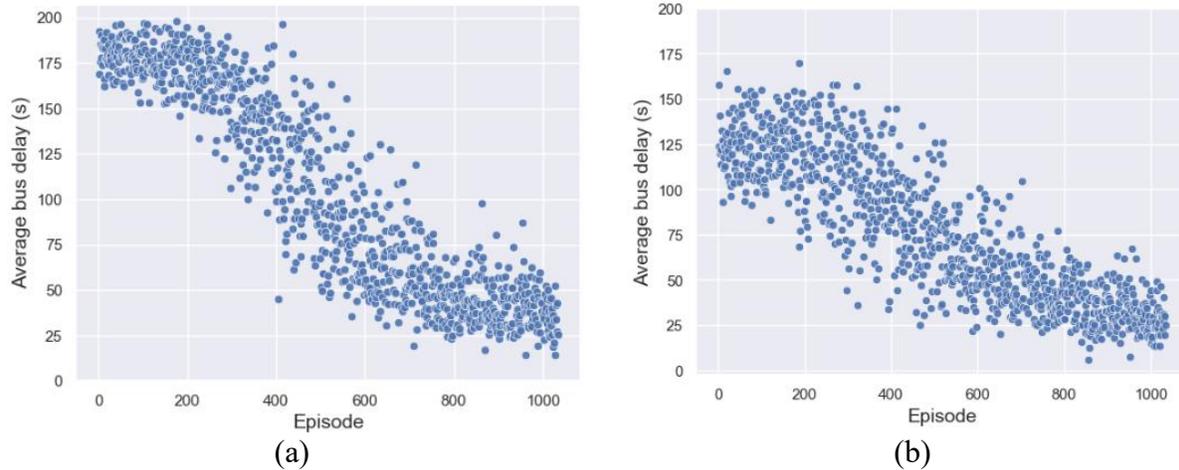

|  |  |
|---|---|
| (a) | (b) |

Figure 7: Average bus delay at (a) Intersection A and (b) Intersection B during training for coordinated TSP agents

**(b) Bus travel time**

Figure 8 shows bus travel with and without TSP for coordinated TSP agents. The setup for test runs and the plotted movements are the same as described for the independent TSP agents in section 4.2 and Figure 5. Across the two intersections (i.e., Inter A&B_EB), coordinated TSP agents achieve a higher reduction in bus travel time compared to independent TSP agents. Bus travel time reduces by 27% compared to the 22% achieved with independent TSP agents. Contributing factors for this improvement may include: (1) the utilization of coordinated TSP strategy reward functions specifically focused on multi-intersections operations, (2) intersection B having additional lead-time to adjust the signal timing when the bus is detected at A, and (3) improved stability in the training.

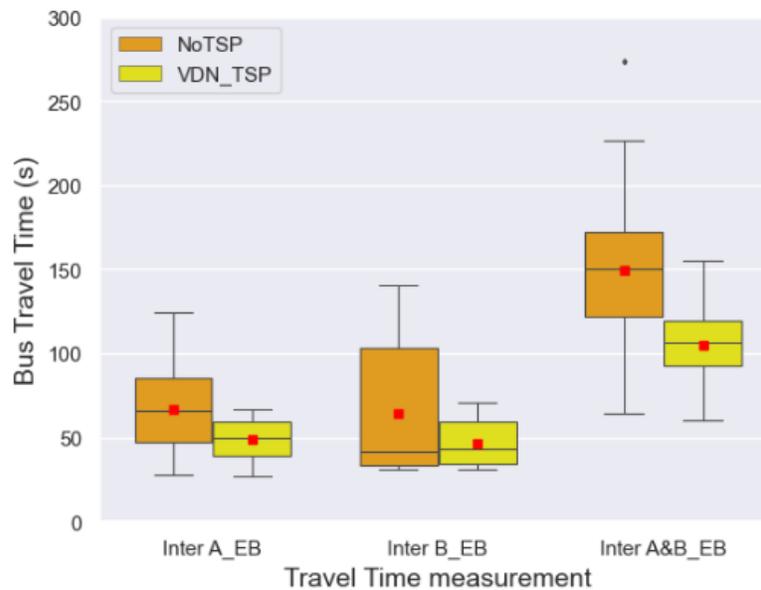

Figure 8:Bus travel time with and without TSP for coordinated TSP agents

**(c) Side street Delay**

Figure 9 shows general traffic delay for the selected side street movements with and without TSP for coordinated TSP agents. For all movements except for A_SB_LT, side street delay increases slightly after TSP. The unexpected slight reduction in delay for A_SB_LT is likely due to the algorithm "overcompensating" for the time taken from the movement after TSP.

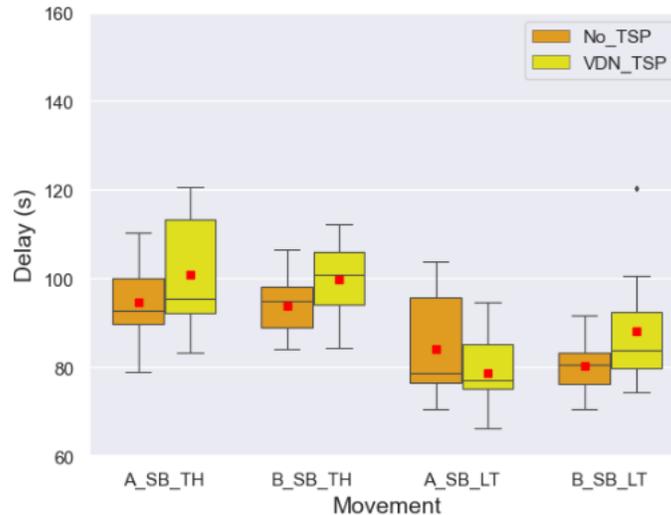

Figure 9: Side Street delay with and without TSP

## 5   CONCLUSIONS AND RECOMMENDATIONS

This study integrated TSP into MARL for signal control. The first part of the study developed adaptive signal control based on the MARL and CTDE paradigm for a pair of coordinated intersections in a microscopic simulation environment. The two agents, one for each intersection, were centrally trained under a VDN architecture. The objective was to minimize total network delay while achieving signal coordination in a prioritized direction. The trained agents show slightly better performance compared to coordinated actuated signal control based on overall intersection delay at v/c of 0.95. In the second part of the study event based TSP is developed. In one variation independent TSP agents are formulated and trained under a DTDE framework to implement TSP at each intersection when a bus is present. In the second variation, the two TSP agents are centrally trained under CTDE framework and VDN architecture to select and implement coordinated TSP strategies across the two intersections when a bus is present. In both cases the agents converge to the same bus delay value, but independent agents show high instability throughout the training process. The two independent agents reduce bus delay across the two intersections by 22% compared to the no TSP case while the coordinated TSP agents achieve 27% delay reduction. In both cases, there is a slight increase in delay within five minutes of bus check-in for the majority of bus side-street movements.

In future studies, the formulated algorithms will be tested on a larger corridor model with more intersections, actual volumes, different bus stop locations, relaxed constraints on signal phasing, and additional constraints such as pedestrian provisions. This will enable the evaluation of the scalability and robustness of the developed algorithms.